# Privacy-Preserving Machine Learning for Electronic Health Records using Federated Learning and Differential Privacy

Naif Ganadily & Han J. Xia
*University of Washington, Department of Electrical and Computer Engineering*

*Abstract* — An Electronic Health Record (EHR) is an electronic database used by healthcare providers to store patients' medical records which may include diagnoses, treatments, costs, and other personal information. Machine learning (ML) algorithms can be used to extract and analyze patient data to improve patient care. Patient records contain highly sensitive information, such as social security numbers (SSNs) and residential addresses, which introduces a need to apply privacy-preserving techniques for these ML models using federated learning and differential privacy.

## I. Introduction

With the increased application of machine learning (ML) in the healthcare industry to diagnose patients or to prescribe medication, applying a privacy-preserving machine learning (PPML) framework for Electronic Health Record (EHR) systems allows healthcare providers to collaboratively train and evaluate ML models without exposing sensitive patient records. EHRs are prone to attacks by cybercriminals, hackers, unauthorized third parties, and even administrators operating within the healthcare system. Attacks may occur in the form of data breaches, insecure communication channels, or insufficient access controls. Given the sensitive nature of patient records, these risks must be mitigated to protect the trust between patients and those working in the healthcare industry, and to protect healthcare providers against potential legal repercussions in the event of an unauthorized access.

PPML algorithms are also needed to comply with federal regulations, such as the Health Insurance Portability and Accountability Act (HIPAA) in the United States or the General Data Protection Regulation (GDPR) in the European Union. These privacy laws were created to protect personally identifiable information from being disclosed without their knowledge or consent.

## II. Architecture

The EHR architecture described in this section is based on the literature, Aspects of privacy for electronic health records (Haas, et al.). As shown in Fig. 1, the architecture consists of two subsystems – the patient services scheme and the data services scheme. The patient services subsystem focuses on administrative communication and provides an interface for patients to express their consent and agreement for privacy policies regarding the usage of their data. The data service controls the flow and storage of medical records and provides an interface for different medical providers to access its data in a secure manner.

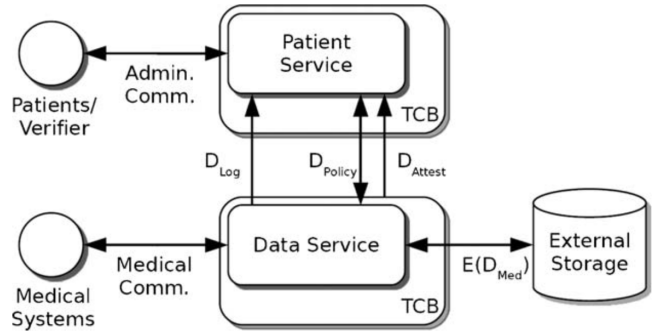

Fig. 1    EHR architecture overview [1].

The patient service scheme shown in Fig. 2 has three primary functions: (1) to provide a means of communication with patients, (2) to control the flow of information, and (3) to allow verifiers to ensure that all agreed-upon policies have been enforced. The policy management allows patients to view, express their consent, and to modify privacy policies that describe their access rights. The logging service uses hash chains to store authentic log files for every access request generated from the policy service. When the verification service obtains log files from the logging service, it uses this information to prove to the verifier that the system has enforced the agreed-upon policy.

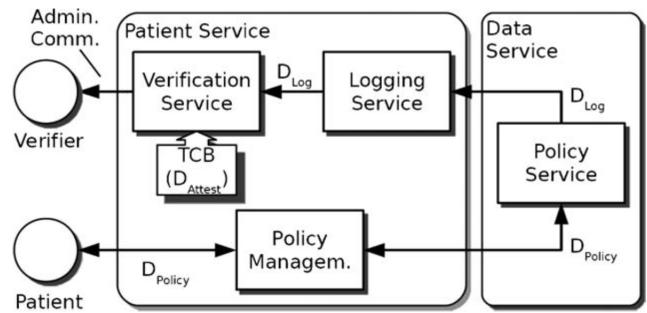

Fig. 2    Scheme of the patient service [1].

The purpose of the policy service within the data service subsystem, as shown in Fig. 3, is to inform the policy enforcement point (PEP) the validity of an access request based on the previously agreed-upon privacy policy. The policy service also sends the resulting decision to the patient service for logging. The watermarking scheme (WMS) applies an asymmetrical digital watermarking (or fingerprinting) scheme, wherein the provider and the consumer receive different watermarks. When one of the parties has violated the agreed-upon policy, the patient is able to determine the source of the leak. Before any medical data is transferred to the external storage, the pseudonymity service (PS) encrypts it with a key that is unique to every patient. Upon a pull request from the external storage, the PS decrypts it and the WMS applies a digital watermark before it is sent to the requesting party with the respective obligations.

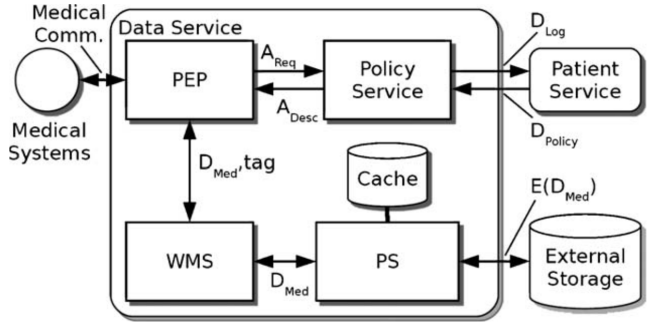

Fig. 3    Scheme of the data service.

The diagram in Fig. 4 illustrates the actual implementation of the EHR in Python, which describes the relationships between elements. For instance, the hospital has doctors, treats patients, and generates records. Records belong to patients and allows one to generate health risk assessments, which can be performed by the doctor. The logger can log all transactions involving the doctor, patient, records, and pharmacies. The role element manages the user visibility of the log based on their status involving the EHR. This information may be only accessed using a key, which contains different privileges for different actors. For instance, a patient's social security number (SSN) and their charges may only be visible to a healthcare insurer whereas a physician is prohibited from viewing a patient's SSN and medical charges.

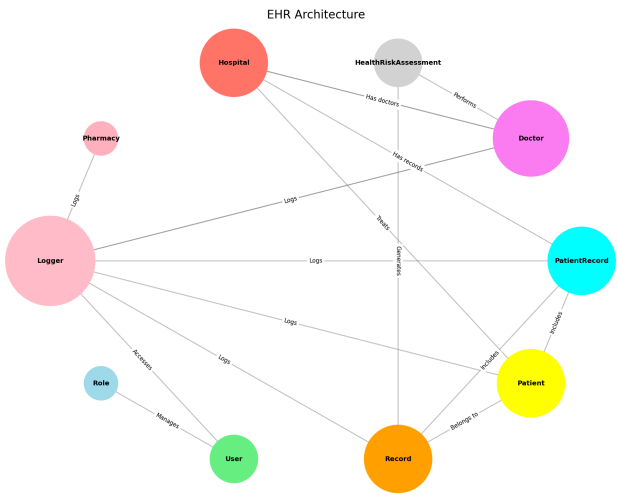

Fig. 4    Actual implementation of the EHR architecture.

### III. DATA EXTRACTION AND MODELLING

To build a sample ML model, the data was extracted from two different sources. The first dataset was obtained from Synthea, which is an open-source, synthetic patient generator that models the medical history of synthetic patients and provides a realistic EHR dataset for research purposes [2]. The second source was extracted from Kaggle for medical cost personal datasheets, which contains information such as age, sex, BMI, number of children, smoker status, and cost of charges [3]. To model a realistic dataset, the SSNs were extracted from the Synthea synthetic patient records and merged with the Kaggle medical cost personal datasets, as demonstrated in Fig. 5.

|  | age | sex | bmi | children | smoker | region | charges | SSN |
|---|---|---|---|---|---|---|---|---|
| 0 | 45 | female | 25.175 | 2 | no | northeast | 9095.06825 | 999-76-6866 |
| 1 | 36 | female | 30.020 | 0 | no | northwest | 5272.17580 | 999-73-5361 |
| 2 | 64 | female | 26.885 | 0 | yes | northwest | 29330.98315 | 999-27-3385 |
| 3 | 46 | male | 25.745 | 3 | no | northwest | 9301.89355 | 999-85-4926 |
| 4 | 19 | male | 31.920 | 0 | yes | northwest | 33750.29180 | 999-60-7372 |
| ... | ... | ... | ... | ... | ... | ... | ... | ... |
| 1166 | 32 | female | 20.520 | 0 | no | northeast | 4544.23480 | 999-60-9291 |
| 1167 | 35 | female | 35.815 | 1 | no | northwest | 5630.45785 | 999-29-3501 |
| 1168 | 44 | male | 22.135 | 2 | no | northeast | 8302.53565 | 999-60-2184 |
| 1169 | 49 | female | 23.845 | 3 | yes | northeast | 24106.91255 | 999-48-3257 |
| 1170 | 24 | female | 23.210 | 0 | no | southeast | 25081.76784 | 999-72-8988 |

1171 rows × 8 columns

Fig. 5    List of patients containing charges, SSN, and other information.

The combined data was used to establish a linear regression model using the LinearRegression library from sklearn.linear_model to predict a patient's charges based on their sex, smoker status, and region. The model's R-squared value of 0.76 indicated a relatively strong association between the selected categories and charges. From these results, there were opportunities to improve the model as the root mean squared error (RMSE) value indicates that the difference between the predicted and actual charges was about $5,620.58. As demonstrated in Table 1, The LazyPredict library provided the tools to automate multiple regression models and to evaluate their parameters. The GradientBoostingRegressor had the highest R-squared value of 0.86 and an RMSE of 4224.49, which is a relatively stronger fit and a smaller prediction error compared to that of the linear regression model.

Table 1    Regression models generated with LazyPredict.

| Model | R-Squared | RMSE |
|---|---|---|
| **GradientBoostingRegressor** | 0.86 | 4224.49 |
| **LGBMRegressor** | 0.86 | 4345.92 |
| **HistGradientBoostingRegressor** | 0.85 | 4421.81 |
| **RandomForestRegressor** | 0.85 | 4502.75 |
| **XGBRegressor** | 0.84 | 4661.15 |
| **BaggingRegressor** | 0.83 | 4761.82 |
| **AdaBoostRegressor** | 0.82 | 4878.86 |
| **ExtraTreesRegressor** | 0.79 | 5268.94 |
| **KNeighborsRegressor** | 0.78 | 5374.13 |
| **PoissonRegressor** | 0.77 | 5546.43 |
| **LarsCV** | 0.76 | 5620.58 |
| **Lars** | 0.76 | 5620.58 |
| **LassoLarsCV** | 0.76 | 5620.58 |
| **TransformedTargetRegressor** | 0.76 | 5620.58 |
| **LinearRegression** | 0.76 | 5620.58 |
| **Ridge** | 0.76 | 5620.69 |
| **RidgeCV** | 0.76 | 5620.69 |
| **LassoLars** | 0.76 | 5620.84 |
| **Lasso** | 0.76 | 5620.84 |
| **BayesianRidge** | 0.76 | 5620.94 |
| LassoCV | 0.76 | 5623.21 |
| SGDRegressor | 0.76 | 5625.24 |
| OrthogonalMatchingPursuitCV | 0.76 | 5644.55 |
| LassoLarsIC | 0.76 | 5650.11 |
| ExtraTreeRegressor | 0.68 | 6454.67 |
| ElasticNet | 0.68 | 6544.39 |
| HuberRegressor | 0.65 | 6747.52 |
| PassiveAggressiveRegressor | 0.65 | 6772.16 |
| OrthogonalMatchingPursuit | 0.62 | 7082.67 |
| DecisionTreeRegressor | 0.61 | 7167.37 |
| RANSACRegressor | 0.58 | 7420.83 |
| TweedieRegressor | 0.57 | 7524.46 |
| GammaRegressor | 0.5 | 8095.7 |
| ElasticNetCV | 0.14 | 10670.56 |
| DummyRegressor | 0 | 11501.77 |
| NuSVR | -0.02 | 11588.4 |
| SVR | -0.08 | 11951.69 |
| QuantileRegressor | -0.09 | 11973.64 |
| KernelRidge | -0.55 | 14298.72 |
| LinearSVR | -1.08 | 16572.8 |
| MLPRegressor | -1.18 | 16956.33 |
| GaussianProcessRegressor | -457.67 | 245956.84 |

## IV. DIFFERENTIAL PRIVACY

Differential privacy (DP) is a technique used to limit an external user's influence on the outcome of a computation, such as the parameters of an ML model. To simulate DP, noise was added to the 'charges' column of the patient records described in Section III. The highlighted columns in Fig. 6 compare the medical charges before adding noise (green) and after (magenta).

Fig. 6  Comparison of charges with and without DP applied.

The plot in Fig. 7 was generated to aid in visualizing the effects of applying DP to the dataset. The blue circles represent the original charges and the red X's are the result of added noise.

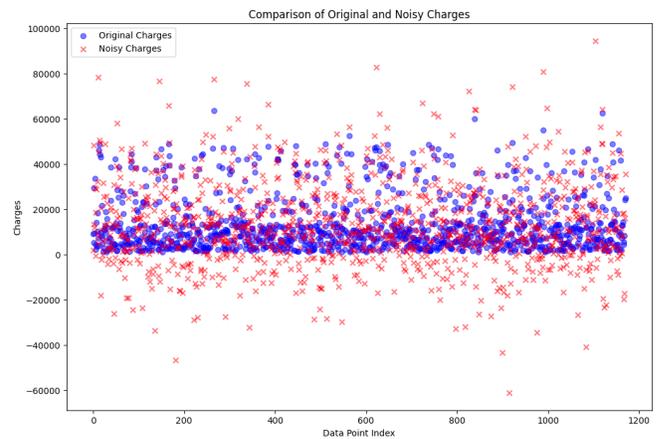

Fig. 7  Visual representation of the charges with added noise.

## V. PSEUDONYMIZATION

Pseudonymization is an encryption technique that replaces personally identifiable information with meaningless information so that it cannot be attributed to a particular identity. As SSN is considered a highly sensitive component of a person's identity, a pseudonymization scheme was applied using a secure hash function. The hashlib library provide the functionality needed to replace each patients' SSN with a SHA-256 pseudonym, which is a 256-bit unreadable value. As shown in Fig. 8, this provides protection against leaks or malicious use of the patient data because each SSN cannot be decrypted without a unique key that only the EHR stores.

Fig. 8  Patient SSNs before pseudonymization (left) and after (right).

## VI. FEDERATED LEARNING

Federated learning (FL) is a ML setting where multiple devices (clients) collaboratively learn a shared model while keeping all the training data on the original device and decoupling the ability to perform ML from the need to store the data in the cloud. This has the advantage of privacy by design, where the raw data is never exposed to the server, and sensitive information remains on the device.

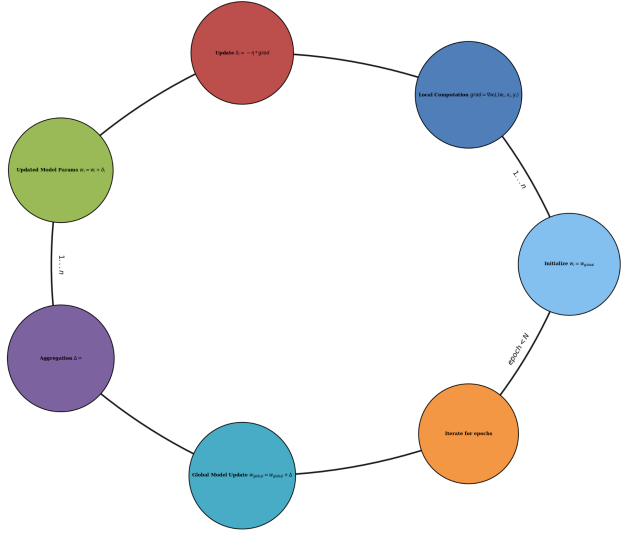

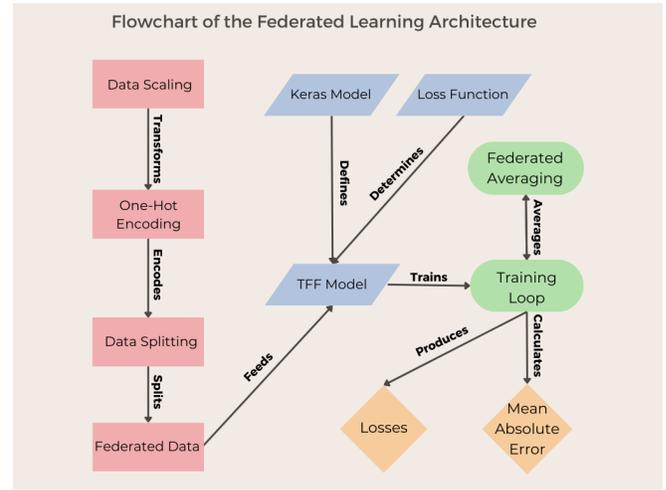

Fig. 9     General flow of the FL process.

Fig. 10     Flowchart of the federated learning process.

The general flow of the Federated Learning process [Fig. 9] can be described as follows:

1. **Initialization**: The server initializes the global model parameters $w$.
2. **Broadcast**: The server sends these parameters $w$ to all participating devices (clients).
3. **Local Computation**: Each client computes an update to the model parameters based on its local data. Specifically, the client computes a gradient of the loss function with respect to the model parameters. Let $x_i$ represent the local data for client $i$ and let $y_i$ represent the labels. The client computes a local update $\delta i$ as follows:

$$\delta_i = -\eta \nabla_w L(w; x_i, y_i)$$

where $L$ is the loss function, $\eta$ is the learning rate, and $\nabla$ denotes the gradient.

4. **Send Model Updates**: Each client sends its computed update $\delta i$ back to the server.
5. **Aggregation**: The server aggregates the updates from each client to compute an overall update. The simplest way to do this is to compute an average:

$$\Delta = \frac{1}{n}\sum_{i=1}^{n} \delta_i$$

where $n$ is the total number of clients in our case 3.

6. **Global Model Update**: The server updates the global model parameters based on the aggregated update:

$$w = w + \Delta$$

7. **Iterate**: The process repeats from step 2 until convergence, i.e., until the change in the global model parameters is smaller than a specified threshold.

The implementation order of the federated learning process in Python can be described as follows [Fig. 10]:

1. Data Preparation: The script begins by manipulating the dataset. Numerical columns ('age', 'bmi', 'children', and 'charges') are standardized using sklearn's StandardScaler. This ensures that these features have a mean value of 0 and a standard deviation of 1, reducing the sensitivity of the model to varying scales. The script also converts categorical columns ('sex', 'smoker', and 'region') into binary vectors via one-hot encoding. An unneeded column, 'SSN', is subsequently discarded.
2. Data Segregation: The processed data is then divided into three segments, each intended for a different participant ('Naif', 'Han', 'Tamara'). This aligns with the concept of federated learning, where each participant has a unique subset of the overall data.
3. TensorFlow Dataset Creation: The script transforms each client's data into a TensorFlow dataset, which is then batched. The model's inputs are all columns excluding 'charges', while 'charges' is set as the target variable.
4. Formation of Federated Dataset: All the individual client datasets are gathered into a list to form a federated dataset.
5. Model Architecture: Using Keras, the script defines a neural network model with two dense layers (using ReLU activation) and a final output layer (without any activation function), typically used in regression problems.
6. TFF Model Initialization: The Keras model is wrapped into a TFF model with the help of the tff.learning.from_keras_model function [5]. The model uses mean absolute error (MAE) as its loss function and mean squared error (MSE) as its performance metric.
7. Setting up Federated Learning Process: The script sets up the federated learning process by calling the tff.learning.build_federated_averaging_process

function. This function establishes a federated averaging process, enabling the model to learn from the federated data. The client optimizer chosen for this process is the stochastic gradient descent (SGD) with a learning rate of 0.01.

8. Model Training: The model training phase consists of a certain number of rounds. In each round, the script calls the 'next' method of the federated averaging process. This method executes one federated averaging step, which involves training the model on each client's data, transmitting the model updates to the server, and averaging these updates.
9. Model Evaluation: After each round of training, the script logs the model's loss and MAE. These values are subsequently visualized to demonstrate how the model's performance evolves over the training rounds.

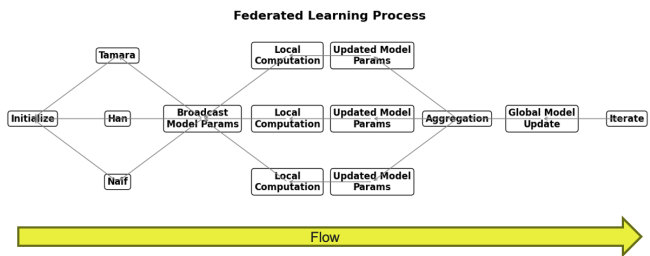

Fig. 11    Simplified FL process as implemented in Python.

The diagram in Fig. 11 summarizes the FL process from a high-level perspective.

## VII.    RESULTS

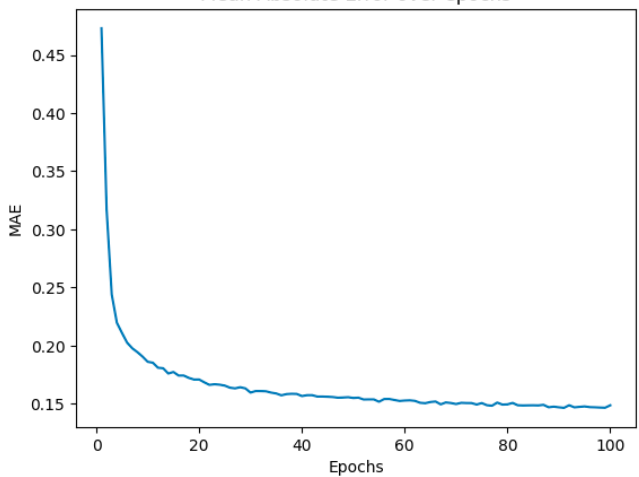

Fig. 12    Mean absolute error vs. epochs.

The Mean Absolute Error (MAE) measures errors between paired observations expressing the same phenomenon. MAE is commonly used in ML as a standard metric to quantify loss in regression problems and is calculated as the average absolute differences between the predicted and actual values. The plot in Fig. 12 illustrates how the MAE has an inversely proportional relationship with respect to the number of epochs. It can be observed that the loss begins to plateau around 80-90 epochs.

## VIII.    CONCLUSION

The results of this study demonstrate the importance of applying privacy-preserving methods for ML models used on medical records stored in EHRs. It is worth noting that the implementation in this study assumed that all clients were available for each round of training, which may be different in a realistic federated learning scenario. For practical purposes, one should implement client selection strategies to select a subset of clients for each training round. Each client also had identical weights for the model update but for realistic modeling, weights should have variation amongst clients based on factors such as the amount of local data, the quality of local data, or the client's contribution to the global model's performance. Secure aggregation is also an essential aspect of federated learning which allows the server to aggregate model updates from clients without being able to inspect individual updates, thereby enhancing privacy. covered in our current implementation. DP is another complementary technique in FL that helps preserve the privacy of the client's data. Future improvements could include mechanisms for adding noise to model updates to ensure DP, which was already done as a standalone. Moreover, model compression tools like Tensorlite could be considered to reduce the cost of communication in FL environments. Lastly, the current implementation aims for a global model that performs well on all clients. In some scenarios, allowing for local model adaptations for personalization could be beneficial.